\documentclass{article}

\usepackage[preprint]{neurips_2025}

\usepackage[english]{babel}

\usepackage{amsmath}
\usepackage{amsfonts}
\usepackage{graphicx}
\usepackage[colorlinks=true, allcolors=blue]{hyperref}
\usepackage{xspace}
\usepackage{booktabs}
\usepackage{tikz}
\usetikzlibrary{positioning}
\usepackage{multirow, makecell}
\usepackage{accents}
\usepackage{siunitx}
\newlength{\dhatheight}

\newcommand{\mnamens}{DVP+} 
\newcommand{\mname}{\mnamens\xspace}
\newcommand\norm[1]{\left\lVert#1\right\rVert}
\newcommand{\down}{$\downarrow$}
\newcommand{\up}{$\uparrow$}

\DeclareMathOperator{\sigmoid}{sigmoid}

\title{
Generative Learning of Differentiable Object Models for Compositional Interpretation of Complex Scenes}

\author{%
  Antoni Nowinowski \\
  Institute of Computing Science\\
  Poznan University of Technology\\
  Poznan, Poland\\
  \texttt{antoni.nowinowski@doctorate.put.poznan.pl}
  \And
  Krzysztof Krawiec \\
  Institute of Computing Science\\
  Poznan University of Technology\\
  Poznan, Poland\\
\texttt{krzysztof.krawiec@cs.put.poznan.pl}
}

\begin{document}
\maketitle

\begin{abstract}

This study builds on the architecture of the Disentangler of Visual Priors (DVP), a type of autoencoder that learns to interpret scenes by decomposing the perceived objects into independent visual aspects of shape, size, orientation, and color appearance. These aspects are expressed as latent parameters which control a differentiable renderer that performs image reconstruction, so that the model can be trained end-to-end with gradient using reconstruction loss. In this study, we extend the original DVP so that it can handle multiple objects in a scene. We also exploit the  interpretability of its latent by using the decoder to sample additional training examples and devising alternative training modes that rely on loss functions defined not only in the image space, but also in the latent space. This significantly facilitates training, which is otherwise challenging due to the presence of extensive plateaus in the image-space reconstruction loss. To examine the performance of this approach, we propose a new benchmark featuring multiple 2D objects, which subsumes the previously proposed Multi-dSprites dataset while being more parameterizable. We compare the DVP extended in these ways with two baselines (MONet and LIVE) and demonstrate its superiority in terms of reconstruction quality and capacity to decompose overlapping objects. We also analyze the gradients induced by the considered loss functions, explain how they impact the efficacy of training, and discuss the limitations of differentiable rendering in autoencoders and the ways in which they can be addressed.   
\end{abstract}

\section{Introduction}\label{sec:intro}

Building correct models of reality requires identifying the causes of observed phenomena and disentangling their interactions. In computer vision (CV), the characteristics of a pixel is determined by a multitude of \emph{aspects} (shapes, positions, and orientations of objects, their color characteristics, lighting, etc.), which interact in complex ways governed by the physical laws. The majority of contemporary CV approaches assume that the generic, connectionist substrate of deep learning (DL) is sufficient to learn these causal chains and form the necessary representations. While multiple successful deployments of CV systems seem to support this claim, most of them do not rely on a principled model of image formation, and, as a result, struggle to truly understand scene content.  Even if good enough most of the time, such models ultimately fail, often in spectacular ways and in unexpected `failure modes', showing puzzling susceptibility to confounding factors/variables, spurious correlations, and outliers (see, e.g., \cite{Zech_2018, DeGrave_2021}). 

Among recent works that attempt to address this deficiency (reviewed in Sec.\ \ref{sec:related}), the DVP proposed in \cite{10.1007/978-3-031-71167-1_13} 
 is a DL architecture that disentangles the visual aspects of perceived objects and then combines them to reconstruct the scene. 
It comprises a convolutional encoder, which delineates an object from the scene and captures its characteristics in terms of color, shape, and other aspects, with a decoder that attempts to reconstruct the scene by rendering a possibly similar object. Thanks to differentiable rendering, DVP can be trained end-to-end with a gradient on raw image data. 

In this study, we propose an approach based on similar principles and bring in the following contributions. 
(i) We design a significantly generalized architecture, dubbed \mname, which can handle multiple objects (Fig.\ \ref{fig:diagram}, Sec.\ \ref{sec:arch}). (ii) We use the renderer as scene generator and use it for alternative training modes, addressing the main challenge of differentiable rendering, i.e. the abundance of plateaus in the reconstruction loss, which leads to difficulties in training (Sec.\ \ref{sec:training}). (iii) Last but not least, we devise a method for acquiring shape prototypes that does not require training of the entire model and streamlines its preparation (Sec.\ \ref{sec:training}). (iv) We demonstrate these advantages with experiments that involve a new benchmark (Sec.\ \ref{sec:experiment}) and explain the differences between training modes with ablations and analysis of gradients.

\section{\mname: The Disentangler of Visual Priors}\label{sec:arch}

\begin{figure}[t]
    \centering
    \includegraphics[width=1.0\textwidth]{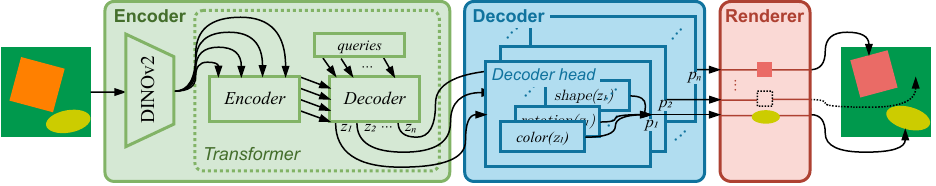}
    \caption{\mname  reconstructs the scene by structurally modeling each object and rendering it. }
    \label{fig:diagram}
\end{figure}

\mname attempts to reconstruct the observed image in a nontrivial fashion. In conventional autoencoders, this is achieved by forcing the model to encode the image in a low-dimensional latent representation. \mname constraints the reconstruction process by explicitly implementing image formation, starting from delineating individual objects, encoding their shapes and colors, and finally rendering them on a raster. By realizing all stages in a differentiable fashion, it sustains the capacity of end-to-end training with gradient-based algorithms. 
In the following, we detail the core components of \mname (Fig.\ \ref{fig:diagram}): the \textbf{encoder} responsible for image analysis and delineation of individual objects, the \textbf{decoder head} that predicts and disentangles the visual aspects characterizing each object, and the \textbf{renderer} that paints the reconstructed objects on an output raster canvas.    

The \textbf{encoder} is based on the Detection Transformer (DETR) blueprint \cite{detr}, which we choose due to its capacity of parsing visual scenes into separate objects. It comprises a perception module followed by an encoder-decoder transformer (EDT). The former is DINOv2 \cite{oquab2023dinov2}, a pre-trained vision transformer (ViT) acting as a feature extractor. The ViT is queried on the input image $x$ and produces spatially localized latent vectors, which are then augmented with a positional encoding, flattened into a sequence of tokens and passed to the EDT. In the EDT, the intermediate results are processed through cross-attention layers of the decoder, which uses independent `object queries' that learn to index the objects in the image. Each object query is a trainable vector that evokes from EDT a latent vector $z_i \in Z$ describing the $i$-th \emph{object candidate} in the input image; see \cite{detr} for more details on the DETR architecture. As the ordering of $z_i$s is arbitrary, they together form a set $\{z_i\}$ that is subject to subsequent processing. 

Next, each object candidate $z_i$ is processed in parallel by the \textbf{decoder head}, which comprises seven MLPs, each mapping the $z_i$ to interpretable \emph{object parameters} that control separate aspects of the process of image reconstruction/formation:  
\begin{itemize}
    \item Color: $Z \rightarrow \mathbb{R}^3$ (represented as an RGB tuple)
    \item Translation: $Z \rightarrow \mathbb{R}^2$  
    \item Scaling: $Z \rightarrow \mathbb{R}$  
    \item Rotation angle: $Z \rightarrow \mathbb{R}^2$ (represented as the lengths of opponent side and adjacent side)
    \item Shape: $Z \rightarrow [0,1]^m$ (where $m$ is the size of the bank of shape prototypes)  
    \item Confidence: $Z \rightarrow \mathbb{R}$ (probability of the candidate being a part of the scene)
    \item Background color: $Z \rightarrow \mathbb{R}^3$ 
\end{itemize}
The shape predictor uses the softmax activation in its $m$-dimensional final layer, and the resulting vector is used to softly index a bank of $m$ \emph{shape prototypes}, each stored as a vector of Elliptic Fourier Descriptors (EFD, \cite{kuhl1982}). The elements of the EFD are real and imaginary parts of complex coefficients that form the spectrum which is mapped with inverse Fourier Transform to object's contour represented as complex `time series'; see details  in Sec.\ \ref{sec:efd}. 
In principle, rather than learning to choose a shape from a trainable bank of shape prototypes, one could predict shape directly, i.e. map $z_i$ to the vector of EFDs. Indeed, the authors of DVP \cite{10.1007/978-3-031-71167-1_13} did that, but within our framework it fa red much worse than prototype-based method. More importantly, learning shape prototypes allows reasoning about the scene in qualitative terms and categorization of observed entities (cf. Sec.\ \ref{sec:intro}), which opens the door to alternative training methods, detailed in the final part of Sec.\ \ref{sec:training}. 

The $(m+10)$-dimensional vector $p_i$ of parameters of the $i$th object candidate forms the input to the \textbf{renderer}. The renderer is a parameter-less (non-trainable) algorithm that `paints' the candidates on a common canvas, taking into account blending of overlapping fragments of objects (translucence). The background color is determined by the average of $z_i$s predicted for all candidates i.e. $b(\overline{z_{i}})$. The raster image produced in this way forms the final output of \mname and is differentiable w.r.t the scene parameters, enabling the gradient of to be backpropagated through the rendering process. From the numerous differentiable rendering methods proposed in recent years \cite{softras, ravi2020pytorch3d, DVG, Laine2020diffrast, DIB-R++, Jakob2020DrJit}, we opted for PyTorch3D \cite{ravi2020pytorch3d} for its clean programming interface and computational efficiency. This rendering method implements the traditional computer graphics route of approximating the objects with a mesh\footnotemark{} of triangles, which are then rasterized. 
\footnotetext{While PyTorch3D offers high-level interfaces for several classes of objects, including meshes, point clouds, volumetric representations, and NeRFs, we required greater flexibility. Therefore, we implemented parts of the rendering pipeline ourselves using PyTorch3D's performant low-level functions.}

\section{Training modes}\label{sec:training}

We propose to train \mname in modes that vary in the primary space (used as the starting point for sampling examples)  and the loss function. 

\noindent\textbf{Primary space.} 
The natural mode of \mname's training is the conventional autoassociative learning for image reconstruction, employed also in  \cite{10.1007/978-3-031-71167-1_13}. In this case, the source of data is a training set of images $X$. 

We propose an alternative approach that relies on the renderer as \emph{image generator}, in the spirit of numerous studies that used decoders for that purpose, including the seminal variational autoencoders \cite{Kingma2014} and adversarial autoencoders \cite{makhzani2016adversarialautoencoders}. 
The common challenge in this scenario is that the generator (here: the renderer) requires sampling from a latent distribution, which is in general unknown; in our context, it is the distribution of objects' parameters $p$, denoted by $P$ in the following. To address this issue, we devise a sampling technique that uses a partially trained \emph{preliminary model} as the source of $P$. First, an example $x$ is sampled from the training set $X$. The preliminary model (more specifically, its encoder and decoder) is then queried on $x$ to yield the estimated scene parameters $\{\hat{p_i}\}$. This becomes our sample from $P$, which can be then passed to the renderer in order to generate an image. The preliminary model can be obtained, for instance, by training the architecture in the conventional way, i.e. for image reconstruction. As we will demonstrate in Sec.\ \ref{sec:experiment}, this sampling technique proves very effective even if the preliminary model was trained only for a few epochs.  

\noindent\textbf{Loss function.} \label{sec:loss}
We consider two loss functions, both implementing a kind of reconstruction error. 
    The first one is the \textbf{image-space loss function} $L_x(x, \hat{x})$ defined as pixel-wise error, used commonly in autoencoders, and also in \cite{10.1007/978-3-031-71167-1_13} (we use MAE in experiments). As an alternative, we propose a \textbf{parameter-space loss function}, 
    which operates on encoder's output. It calculates the reconstruction error of the parameters in $\{\hat{p}_i\}$ predicted by the encoder with respect to the true parameters in $\{p_i\}$ -- provided, for instance, by the generator introduced above. As the number of predicted objects, i.e. the size of the $\{\hat{p}_i\}$ set, can be different from the number of actual objects $n$, and the ordering of objects can vary too, we engage the Hungarian algorithm to efficiently determine the optimal assignment of the predicted candidates in $\{\hat{p}_i\}$ to the actual objects in $\{p_i\}$. The algorithm determines the optimal assignment that minimizes the per-object matching cost function $C(p_i, \hat{p}_j)$, defined as weighted sum of square differences of selected parameters of object $i$ and candidate $j$: 
    \begin{equation}
    \label{eq:L_m}
    C(p_i, \hat{p}_j) = \norm{p_i.t - \hat{p}_j.t}_2 + 0.1 \norm{p_i.c - \hat{p}_j.c}_2 + 0.01 \norm{p_i.\mathit{conf} - \hat{p}_j.\mathit{conf}}_2
    \end{equation}
    where $t$, $c$, and $\mathit{conf}$ denote, respectively, the translation vector, the RGB triple of object's color, and the confidence, i.e. the respective parts of the vector of properties (Sec.\ \ref{sec:arch}). Once the optimal assignment has been determined, each of the $n$ matched object-candidate pairs is subject to  
    the actual parameter-space loss function $L_p(p, \hat{p})$, 
    which measures the discrepancy over all parameters of matched objects, using weighted distance functions (tuned empirically to individual parameters in preliminary experiments):
    \begin{equation}
    \label{eq:L_p}
    \begin{aligned}
        L_p(p, \hat{p}) = &\norm{p.b - \hat{p}.b}_2       &&\text{\textit{(background color)}}\\
        +& \sum\nolimits_{i=1}^{n} \Big( \log(\hat{p}_i.\mathit{conf})        &&\text{\textit{(BCE on confidence)}}\\
        &\mspace{44mu} + 5\norm{p_i.t - \hat{p}_i.t}_2   &&\text{\textit{(translation)}}\\
        &\mspace{44mu} + \norm{p_i.c - \hat{p}_i.c}_2    &&\text{\textit{(color)}}\\
        &\mspace{44mu} + \norm{p_i.s - \hat{p}_i.s}_2    &&\text{\textit{(scale)}}\\
        &\mspace{44mu} + \norm{p_i.sh - \hat{p}_i.sh}_2  &&\text{\textit{(shape)}}\\
        &\mspace{44mu} + 0.05(1 - \cos(p_i.a - \hat{p}_i.a) \cdot sym(p_i.sh))^2 \Big) 
                                                           &&\text{\textit{(rotation)}}\\
        +& \sum\nolimits_{j=n+1}^{m} - \log(1 - \hat{p}_j.\mathit{conf})      &&\text{\textit{(BCE on confidence)}}
    \end{aligned}
    \end{equation}    
where $b$, $s$, $sh$ and $a$ are, respectively, the properties of background color, scale, shape and rotation angle. $sym$ is the number of symmetries of the shape prototype, calculated from its EFD (see Sec.\ \ref{sec:L_p} for details).  
The total loss is the sum of $L_p$s for all $n$ matched pairs. 

$L_p$ induces an error surface (over the space of model's parameters) that is different from $L_x$, which may facilitate training. For instance, when the predicted location of an object in a reconstructed image diverges strongly from its actual location in the input image (e.g.\ to the extent that they do not even overlap), $L_x$ will respond with high reconstruction error, but its gradient with respect to model parameters will be often close to zero. However, the gradient of $L_p$ can be still high, because it depends, among others, on the actual differences in objects' coordinates. The usefulness of both loss functions will be assessed empirically in Sec.\ \ref{sec:experiment}. 

\begin{figure}[t]
    \centering
    \includegraphics[width=1.0\linewidth]{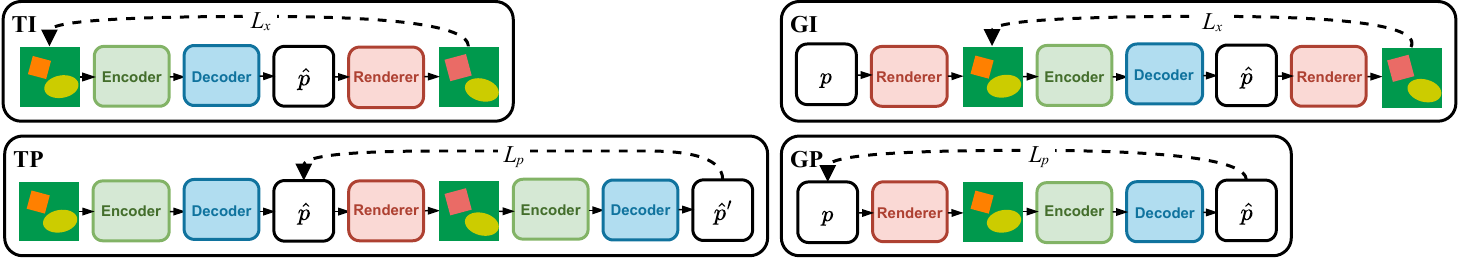}
    \caption{The four training modes of \mname. Top: image-based loss $L_x$, bottom: parameter-space loss $L_p$. Left: images as the source of training data, right: parameters as the source of training data.}
    \label{fig:training-modes}
\end{figure}

\noindent\textbf{Training modes.} 
The choice of the above primary space (\textbf{T}raining set or \textbf{G}enerator) and loss function (\textbf{I}mage space vs. \textbf{P}arameter space) leads to four  \emph{training modes} of \mname (Fig.\ \ref{fig:training-modes}). Let $f$, $g$, and $r$ denote respectively the encoder, the decoder, and the renderer.
\begin{description}
    \item[TI] $\min_{x\sim X} L_x(x, r(g(f(x))))$: The typical way of training autoencoders: the model is queried on an image $x$ from the training set $X$, responds with an image $r(g(f(x))$, and is trained by minimizing the image-space reconstruction loss $L_x$.  
    \item[GP] $\min_{p\sim P} L_p(p, g(f(r(p))))$\footnotemark{}: This mode is dual with respect to TI: a sample $p$ of object parameters is drawn from a given distribution $P$  (e.g. the image generator described above) and rendered. The rendered image is fed into encoder $f$ and decoder $g$ (i.e., $g\circ f$), resulting in predicted parameters of candidates, $g(f(r(x)))$, which are compared to $p$ with parameter-space loss $L_p$. 
    \footnotetext{For the sake of brevity. we abuse the notation by using $L_p$ to denote both the per-object loss (Eq.\ \ref{eq:L_p}) as well the loss summed over all object-candidate pairs.}
    \item[GI] $\min_{p\sim P} L_x(r(p), r(g(f(r(p)))))$: Like GP, but both the sampled parameters $p$ and the predicted parameters $g(f(r(p)))$ are passed through $r$, and the resulting renderings are compared using the image-space loss function $L_x$.  
    \item[TP] $\min_{x\sim X} L_p(g(f(x)), g(f(r(g(f(x))))))$: Like TI, but both the input image $x$ and the reconstructed one $r(g(f(x)))$ are fed into $g \circ f$, and the resulting predictions are compared using the parameter-space reconstruction error $L_p$. 
\end{description}
While the first two modes (TI and GP) are rather obvious, GI and TP map the arguments of the loss function to the other representation space. As argued earlier, we expect this to provide additional informative guidance for the training process.  


\noindent\textbf{Learning of shape prototypes.} 
As signaled earlier, simultaneous optimization of all scene parameters with respect to $L_x$ poses difficulties due to interdependence of their impact on the rendered scene, causing gradients to point to 'misleading' directions, especially when the predicted scene significantly diverges from the target one. As we demonstrate in Sec.\ \ref{sec:experiment}, this happens to be particularly true for shape, as its is strongly entangled with scale and rotation. Indeed, in preliminary experimenting we found out that inaccurate prediction of rotation hinders the emergence of object shapes, and vice versa: learning how to rotate candidates requires having a reasonably accurate model of object's shape. 

To address this challenge in \mname, we use clustering to discover shape prototypes from the training data. Given a sample of training images $X$, for each $x\in X$ we start with obtaining an initial estimate $p$ of scene parameters using a preliminary model (in the simplest case, the same as the one used in the generator described at the beginning of this section). Then, we repeat the following steps. (1) We optimize $p$s using the image-space loss function, i.e. $\min_p L_x(x,r(p))$, obtaining so locally-optimal parameterizations $p^*$. (2) We gather $p^*$s obtained in this way for all $x\in X$ and perform $k$-medoids clustering in the image space\footnotemark{}, in order to capture the typical shapes present in $X$ and discard the outliers. (3) For the medoid $x'$ of each resulting cluster $X'$, we retrieve the corresponding $p'$ and replace with it the $p$s corresponding to the elements of $X'$. After repeating these three steps a number of times, we obtain a stable set of shapes that are representative for $X$, which is then injected into the model as prototype shapes. We found this method to be much more efficient and effective than the one used in DVP, and use it in all experiments reported in this paper.
\footnotetext{We use the MSE between the equally scaled and centered renderings of shapes as the distance metric for clustering. The optimal number of clusters, $k$, is determined using the silhouette score \cite{Rousseeuw1987}.}

\section{Related work}\label{sec:related}

We focus on reviewing past works related to the central topic of this study, i.e. the challenges originating in the characteristics of the loss function. Methods involving differentiable rendering (DR)  frequently face loss function landscapes that simultaneously feature extensive plateaus and are very rugged at places. This issue is present not only when using DR for model training \cite{Im2Vec} (especially in early stages of the training) but also in direct optimization, when the goal is to find the combination of scene parameters that minimize the adopted objective function \cite{DVG, live}. The main reason is that, especially at the beginning of training/optimization, the rendered scene may be completely different from the actual one, with the produced objects failing to overlap with the actual ones. This manifests in low and vanishing gradients and often leads to reaching poor-quality local minima. 

In past studies, this has been addressed in several ways. In the Im2Vec approach \cite{Im2Vec}, the predictions are rendered as image pyramids, with each subsequent level being smaller by a factor 2. The loss is an aggregation of losses at each level, the gradient from the low-resolution levels proving important while processing images with significant dissimilarities. Optimization-based techniques that use explicit primitives, such ad DiffVG \cite{DVG} and Gaussian Splatting (GS) \cite{3Dgaussians}, often start with a large number of primitives (object candidates). This increases the likelihood of some of those candidates overlapping with the target object to be captured, providing a better starting point for optimization. The GS authors further refine the initialization strategy by employing a conventional algorithm (Structure from Motion) to derive the initial parameters of the object candidates. To address the related problem where multiple object candidates might redundantly reconstruct a single object, the LIVE method \cite{live} introduces new object candidates sequentially. This also leads to reconstructing the target with fewer primitives, making the reconstruction more interpretable.
In contrast, volumetric approaches such as Neural Radiance Fields (NeRF) \cite{nerf} model scenes with continuous volumetric scene functions. This formulation, grounded in volume rendering physics, inherently promotes a smoother optimization landscape compared to discrete primitive-based methods, which simplify the optimization process but complicate further processing of the reconstructed scene.

In contrast, we cope with challenging loss functions by engaging the renderer-decoder as scene generator and learning in `dual' modes presented in Sec.\ref{sec:training}. 
In a broader context, \mname and some of the approaches mentioned above, by explicitly modeling visual aspects of objects and rendering them for reconstruction, subscribe to the seminal paradigm of \emph{vision as inverse graphics} \cite{barrow1978recovering}.

\section{Results}\label{sec:experiment}

We focus on comparing the training modes of \mname presented in Sec.\ \ref{sec:training}, i.e. TI, GP, GI, and TP. All configurations of \mname use DINOv2 \cite{oquab2023dinov2} as the pretrained perception model, which remains frozen during training. 

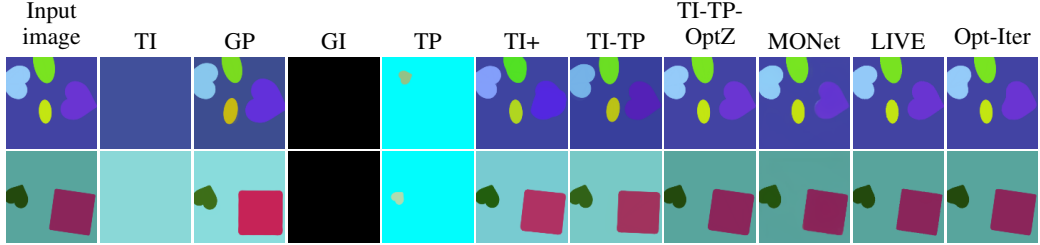
\begin{figure}
    \centering
    \begin{tikzpicture}[
        node distance = 1pt,
        every node/.style = {inner sep=0pt}
    ]
    
      \node (pic1) {\includegraphics[width=0.086\linewidth]{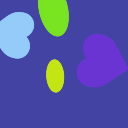}};
      \node[above=1mm of pic1, font=\footnotesize, align=center] {Input \\ image};
      \node[below=1pt of pic1] {\includegraphics[width=0.086\linewidth]{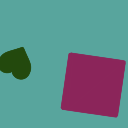}};
    
      \node (pic2) [right=of pic1] {\includegraphics[width=0.086\linewidth]{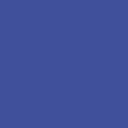}};
      \node[above=1mm of pic2, font=\footnotesize] {TI};
      \node[below=1pt of pic2] {\includegraphics[width=0.086\linewidth]{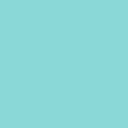}};

      \node (pic3) [right=of pic2] {\includegraphics[width=0.086\linewidth]{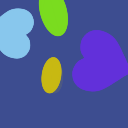}};
      \node[above=1mm of pic3, font=\footnotesize] {GP};
      \node[below=1pt of pic3] {\includegraphics[width=0.086\linewidth]{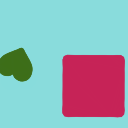}};

      \node (pic4) [right=of pic3] {\includegraphics[width=0.086\linewidth]{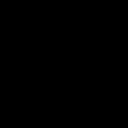}};
      \node[above=1mm of pic4, font=\footnotesize] {GI};
      \node[below=1pt of pic4] {\includegraphics[width=0.086\linewidth]{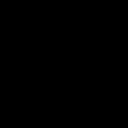}};

      \node (pic5) [right=of pic4] {\includegraphics[width=0.086\linewidth]{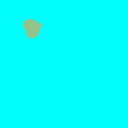}};
      \node[above=1mm of pic5, font=\footnotesize] {TP};
      \node[below=1pt of pic5] {\includegraphics[width=0.086\linewidth]{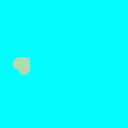}};

      \node (pic6) [right=of pic5] {\includegraphics[width=0.086\linewidth]{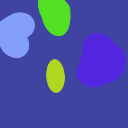}};
      \node[above=1mm of pic6, font=\footnotesize] {TI+};
      \node[below=1pt of pic6] {\includegraphics[width=0.086\linewidth]{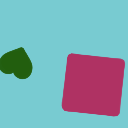}};

      \node (pic7) [right=of pic6] {\includegraphics[width=0.086\linewidth]{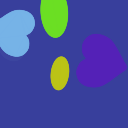}};
      \node[above=1mm of pic7, font=\footnotesize, align=center] {TI-TP};
      \node[below=1pt of pic7] {\includegraphics[width=0.086\linewidth]{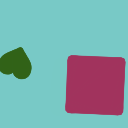}};

      \node (pic7b) [right=of pic7] {\includegraphics[width=0.086\linewidth]{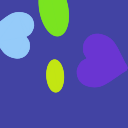}};
      \node[above=1mm of pic7b, font=\footnotesize, align=center] {TI-TP- \\ OptZ};
      \node[below=1pt of pic7b] {\includegraphics[width=0.086\linewidth]{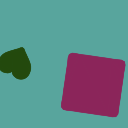}};

      \node (pic8) [right=of pic7b] {\includegraphics[width=0.086\linewidth]{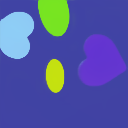}};
      \node[above=1mm of pic8, font=\footnotesize] {MONet};
      \node[below=1pt of pic8] {\includegraphics[width=0.086\linewidth]{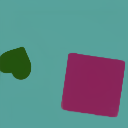}};

      \node (pic9) [right=of pic8] {\includegraphics[width=0.086\linewidth]{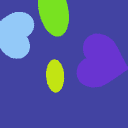}};
      \node[above=1mm of pic9, font=\footnotesize] {LIVE};
      \node[below=1pt of pic9] {\includegraphics[width=0.086\linewidth]{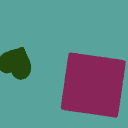}};

      \node (pic10) [right=of pic9] {\includegraphics[width=0.086\linewidth]{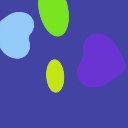}};
      \node[above=1mm of pic10, font=\footnotesize] {Opt-Iter};
      \node[below=1pt of pic10] {\includegraphics[width=0.086\linewidth]{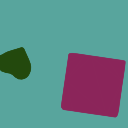}};
    
    \end{tikzpicture}
    \caption{Reconstruction results for selected images from the testing set. }
    \label{fig:mds-hr-ex}
\end{figure}

While many of similar approaches have been in the past evaluated on the Multi-dSprites dataset \cite{Burgess_2019}, we found it lacking in low image dimensions (64 x 64). This inclined us to devise a higher-resolution benchmark, dubbed MDS-HR. The other important difference is that MDS-HR does not feature object occlusion. Using the procedures described there, we generated a training and testing set with similar to Multi-dSprites characteristic, comprising, respectively, \num{55000} and \num{5000} examples, each being a 128 x 128 image featuring from 1 to 4 objects randomly distributed in the image, representing 3 unique shapes (Fig.\ \ref{fig:mds-hr-ex}). Details on MDS-HR and access to the dataset are provided in Sec.\ \ref{sec:mds-hr}; Sec.\ \ref{sec:software} presents our software implementation and provides open access to it.


\textbf{Simple training regimes.}
We start with the `monolithic' training regimes that optimize from scratch \emph{all} trainable parameters with the loss function defined in a given mode. Each training process comprised 100 epochs of the Adam optimizer \cite{DBLP:journals/corr/KingmaB14} with the learning rate 0.0001 (see Sec.\ref{sec:training-details} for more details). Training of a single \mname model lasted approximately 10 hours on a workstation equipped with RTX 3090, Intel Core i7-11700KF CPU and 32 GB of RAM (see Sec.\ \ref{sec:hardware} for details on the hardware architecture).

\begin{table}[t]
\centering
\caption{Reconstruction quality of \mname trained with simple training modes. Best results in bold. }
\label{tab:accuracy}
\begin{tabular}{lrrrrrr}
\toprule
Method & MAE \down & MSE \down & SSIM \up & IoU \up & ARI \up \\
\midrule
\mname{TI} & 0.1115 & 0.0404 & 0.8154 & 0.0870 & 0.0    \\
\mname{GP} & \textbf{0.0637} & \textbf{0.0107} & \textbf{0.8540} & \textbf{0.8452} & \textbf{0.8235} \\
\mname{GI} & 0.4964 & 0.3287 & 0.0202 & 0.0    & 0.0    \\
\mname{TP} & 0.4895 & 0.3216 & 0.4224 & 0.0653 & 0.0838 \\
\bottomrule
\end{tabular}
\end{table}

The generator introduced at the beginning of Sec.\ \ref{sec:training} and used in GP and GI offers  an unfair advantage for \mname, as it exposes the learner to examples from outside the training set -- possibly including those from the test set. Therefore, for GP and GI, we sample only the $p$s that correspond to actual examples from the training set, i.e. $p \sim \{p: r(p) \in X\}$.

Table \ref{tab:accuracy} reports the test-set reconstruction accuracy in terms of Mean Square Error (MSE), Structural Similarity Measure (SSIM, \cite{Wang2004}), Intersection over Union (IoU), and Adjusted Rand Index (ARI, \cite{Hubert_Arabie_1985}). 
MSE an SSIM are calculated directly from pixels' RGB values (scaled to the $[0,1]$ interval). IoU and ARI are more structural and require the rendered pixels to be unambiguously assigned to an object or to the background (for IoU) or to each distinct object (for ARI). Therefore, we force the models to render scenes as binary masks with white objects on a black background. 

GP clearly fares best, which corroborates that training by optimizing parameter-space loss $L_p$ is advantageous compared to optimizing the conventional image-space reconstruction loss $L_x$ (TI). While TI still achieves adequate results on pixel-based metrics, its performance on the structural metrics is substantially worse. In fact, the TI model has learned to reconstruct only the background color, ignoring the objects. A key issue arises early in the training: the initial object candidates produced by the model exhibit little or no resemblance to the ground truth objects. The image-space loss $L_x$ between such dissimilar images does not provide any relevant information (as discussed in Sec.\ \ref{sec:training}), causing the model to end up in a local minimum achievable via reconstructing only the background (see Sec.\ \ref{sec:extra-results} for visualizations). The more sophisticated training modes (GI and TP) fail severely, producing almost constant outputs, regardless of the input. This occurs because their loss functions lack a component enforcing reconstruction of the original input data.


\textbf{Combined training regimes.}
In the next step, we attempt to improve on the  models identified in Table \ref{tab:accuracy}. While \mname{GP} is clearly superior, it has the disadvantage of requiring ground truth scene parameter (sampling of $p$). To mitigate this dependency, we appoint \mname{TI} as the basis for further investigations. 
To address the absence of informative gradient in TI, we initialize the weights (and the weights of all models used in the following) with those of \mname{GP} from an early (4th) training epoch, so that the scene renderings are good enough for $L_x$ to provide meaningful guidance. We then train this \textbf{TI+} configuration with $L_x$, improving on MAE, MSE and IoU (Table \ref{tab:acc-compound}). However, ARI is notably low,  primarily due to the model reconstructing single objects with multiple candidates. ARI is the only metric here that is sensitive to this undesirable behavior, which is not explicitly penalized in training, and thus models may tend to exercise it, as reconstructing an object with multiple candidates/shapes is often easier than with a single candidate.

\begin{table}[t]
\centering
\caption{Reconstruction accuracy for particular training regimes and the baseline methods. }
\label{tab:acc-compound}
\begin{tabular}{lrrrrrr}
\toprule
Method           & MAE \down & MSE \down & SSIM \up & IoU \up & ARI \up \\
\midrule
\mname{GP}        & 0.0637 & 0.0107 & 0.8540 & 0.8452 & 0.8235 \\
\mname{TI+}       & 0.0444 & 0.0050 & 0.9148 & 0.9341 & 0.2275 \\
\mname{TI-TP}     & 0.0482 & 0.0068 & 0.8878 & 0.8884 & 0.8693 \\
\mname{TI-TP-OptP}  & 0.0023 & 0.0002 & 0.9947 & 0.9831 & 0.9780 \\
\mname{TI-TP-OptZ} & 0.0015 & \textbf{0.0001} & \textbf{0.9961} & \textbf{0.9927} & \textbf{0.9820} \\
\midrule
MONet            & 0.0126 & 0.0013 & 0.9676 & 0.9312 & 0.9137 \\
LIVE             & 0.0098 & 0.0010 & 0.9850 & 0.6767 & 0.6740 \\
Opt-Iter         & \textbf{0.0012} & 0.0002 & 0.9941 & 0.9625 & 0.9414 \\
\bottomrule
\end{tabular}
\end{table}


To tackle this challenge, we combine TI and TP into \textbf{TI-TP}, where training consists in optimizing $L_x + L_p$, i.e. optimizing all parameters of the model with the gradient $\nabla L_x(x, r(g(f(x)))) + \nabla L_p(g(f(x)), g(f(r(g(f(x))))))$ for $x\in X$.\footnote{We verified that these functions vary in similar ranges, so their expected impact on training is comparable.}  While TI-TP turns out to deteriorate most metrics compared to TI+, it substantially outperforms \mname{GP} on ARI -- a valuable outcome, as ARI reflects how well one-to-one relation between target objects and object candidates is captured.

Given the superiority of TI-TP on ARI, we adopt it as the basis for the -Opt configurations, in which the predictions made by the encoder $f$ and decoder $g$ of TI-TP serve as starting points for a gradient-based optimization procedure, which tunes the scene parameters individually for each scene \mname is queried on. We consider two variants that vary in the variables that are subject to optimization: in \textbf{TI-TP-OptP}, we optimize $p$ with $\nabla_p r(p)$, while in \textbf{TI-TP-OptZ}, we optimize $z$ with $\nabla_z r(g(z))$. In both cases, we allow $100$ iterations of the Adam optimizer \cite{DBLP:journals/corr/KingmaB14}. These hybrids of DL and optimization achieve best results across all metrics. Interestingly, following the gradient in the latent space $Z$ proves slightly more effective than optimization in the interpretable space of objects' parameters.

To put these results in the context of prior work, Table\ \ref{tab:acc-compound} presents also the performance of MONet \cite{Burgess_2019}, LIVE \cite{live} (Sec.\ \ref{sec:related}), and Opt-Iter, a purely optimization-based algorithm that that hybridizes LIVE with our shape representation.\footnote{We do not include DVP \cite{10.1007/978-3-031-71167-1_13} in the comparison, as it is by design incapable of handling multiple objects.} Opt-Iter iteratively identifies the image region that is most divergent from the current scene approximation, initializes a new object candidate at that location (sampling its color from the region), and then optimizes all candidates introduced so far to minimize $L_x$. The cycle repeats until a predetermined number of objects is reached. See Sec.\ \ref{sec:baselines} for more details on MONet and Opt-Iter and their parameterization.  While the baseline methods fare overall well, they yield to the best configurations of \mname, particularly on the structural metrics. 

Figure \ref{fig:mds-hr-ex} presents examples of reconstructions obtained with \mname and baseline methods. As all combined variants achieve relatively high metrics, the differences between the renderings they provide are subtle, yet noticeable (e.g., the angles of rotation; the sharpness of corners, especially for the heart shape).  More examples of renderings can be found in Sec.\ \ref{sec:extra-results}.

\begin{figure}[t]
    \centering
    \includegraphics[width=0.57\textwidth]{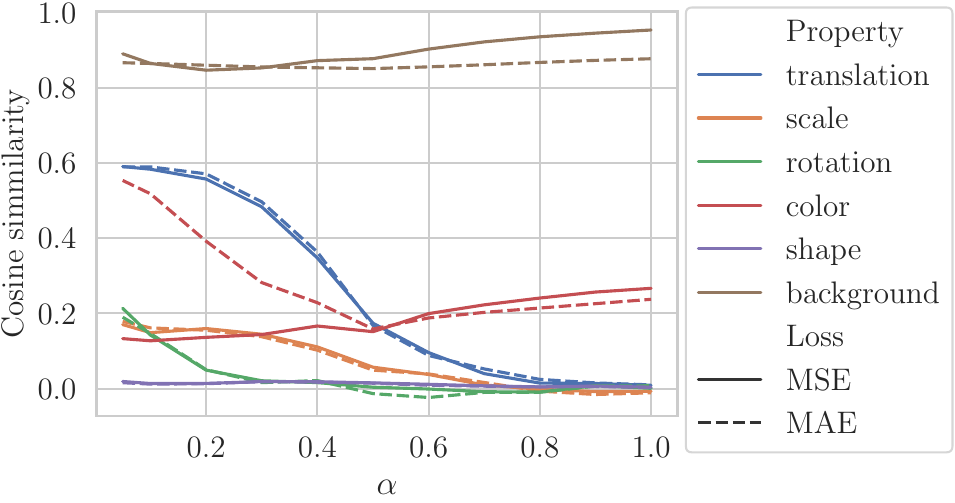}\hfill
    \includegraphics[width=0.43\textwidth, trim=0 6pt 0 0]{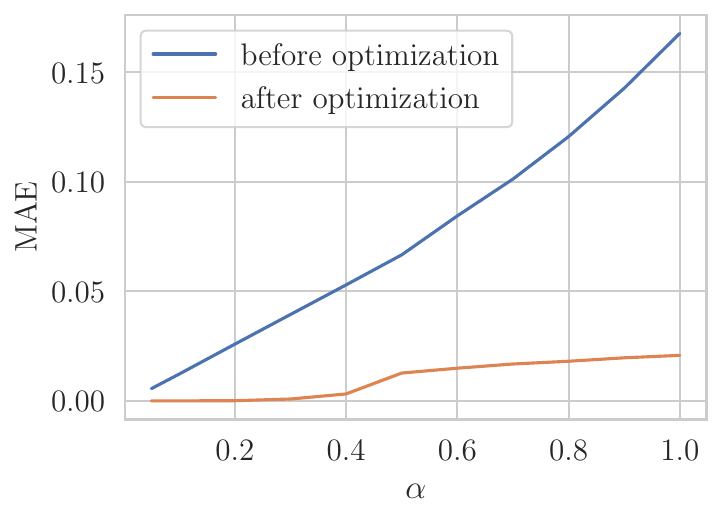}
    \caption{Left: The cosine similarity between gradients calculated of image-space loss and parameter-space loss for perturbed renderings ($\alpha$: perturbation strength). Right: The value of the loss function (MAE) before and after optimization. }
    \label{fig:gradient}
\end{figure}

\textbf{Analysis of gradients.} To provide a better insight into the differences of the learning guidance provided by the image-space loss function $L_x$ and the parameter-space loss function $L_p$, we analyze the renderer $r$ in the following way. We sample a pair of random examples from the training set, retrieve their scene parameter vectors $p_1$ and $p_2$ and combine them linearly $p'=\alpha p_1+(1-\alpha)p_2$. Then, we query the renderer on $p'$, calculate the gradient of $L_x$ function on the resulting scene, $\nabla_{p'} L_x(r(p), r(p'))$, and confront it with the analogous gradient exerted by $L_p$, $\nabla_p L_p(p')$ using the cosine similarity function. Figure \ref{fig:gradient}a presents the similarity as a function of $\alpha$ varying in $[0.1,1]$, averaged over a sample of 2048 pairs of examples, for $p$ factored into parameters responsible for particular scene aspects, separately for MSE- and MAE-based $L_x$. The curves show thus how the direction of the gradient exerted by $L_x$ diverges from that of the `native' gradient $L_p$ when the rendered scene $r(p')$ is perturbed with respect to $p$ (which is meant to represent the true, optimal interpretation of the scene). Except for color properties, the similarity between parameter vectors decreases quickly with the perturbation strength $\alpha$, which explains why training \mname with $L_x$ is more challenging. Color aspects are exempt from this trend, because the RGB components of rendered pixels depend trivially on their representation in $p$; this is particularly visible for the background color. The counterintuitive effect of MSE-based $L_x$ on color arises because MSE heavily focus on small mismatched areas with high color differences, prioritizing them over larger areas of good overlap where color differences are minimal.  Note that the shape parameters exhibit very low similarity, corroborating the difficulty of learning this particular visual aspect. Figure \ref{fig:gradient}b presents the average of $L_p(p')$ and how it can be reduced in 100 steps of optimization with Adam \cite{DBLP:journals/corr/KingmaB14}. The loss can be reduced to near-zero as long as the gradient directions of $\nabla L_x$ and $\nabla L_p$ align.

\textbf{Ablations.}
We ablate the best configurations reported in Table \ref{tab:acc-compound} by eliminating the encoder and/or decoder, in order to assess their usefulness as providers of initial parameter estimates, i.e. starting points for the optimization process. The first  configuration, \textbf{Rand-OptP}, is an ablation of TI-TP-OptP and consists in  optimization \emph{in the parameter space $P$}. We start with randomly initializing the parameters $p$ for $n$ object slots, where $n$ is the number of objects in the target scene. Then we query the renderer $r$ on it, and minimize image loss until convergence by performing descent along the gradient $\nabla_p L_x(x, r(p))$. Note that this configuration does not rely on shape prototypes. 
The second configuration, \textbf{Rand-OptZ}, is an ablation of TI-TP-OptZ, consists in \emph{optimization in the latent space $Z$}. The starting point of the optimization $z$ is randomly drawn from the space of latents $Z$, and the optimization follows $\nabla_{z} L_x(x, r(g(z)))$.  

The metrics achieved by both ablated variants of \mname, presented in Table \ref{tab:ablations}, are much worse than those for the non-ablated TI-TP-OptP and -OptZ models in Table \ref{tab:acc-compound}, clearly indicating that the initial estimates predicted by the model are essential for high-quality scene reconstruction.  

\begin{table}[t]
\centering
\caption{Test-set evaluation of the ablated variants of the \mname-TI-TP-Opt models.}\label{tab:ablations}
\label{tab:}
\begin{tabular}{lrrrrr}
\toprule 
Ablated configuration & MAE \down   & MSE \down   & SSIM \up  & IoU \up   & ARI \up   \\
\midrule
Rand-OptP             & 0.0409 & 0.0178 & 0.8976 & 0.2399 & 0.3046 \\ 
Rand-OptZ            & 0.0502 & 0.0217 & 0.8610 & 0.2532 & 0.1983 \\
\bottomrule
\end{tabular}
\end{table}

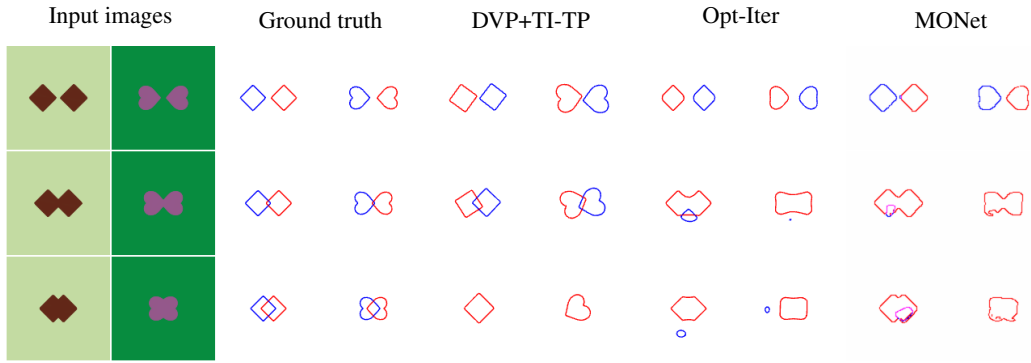
\begin{figure}[t]
    \centering
    \begin{tikzpicture}
        \node (imageNode) at (0, 0) {\includegraphics[width=\textwidth]{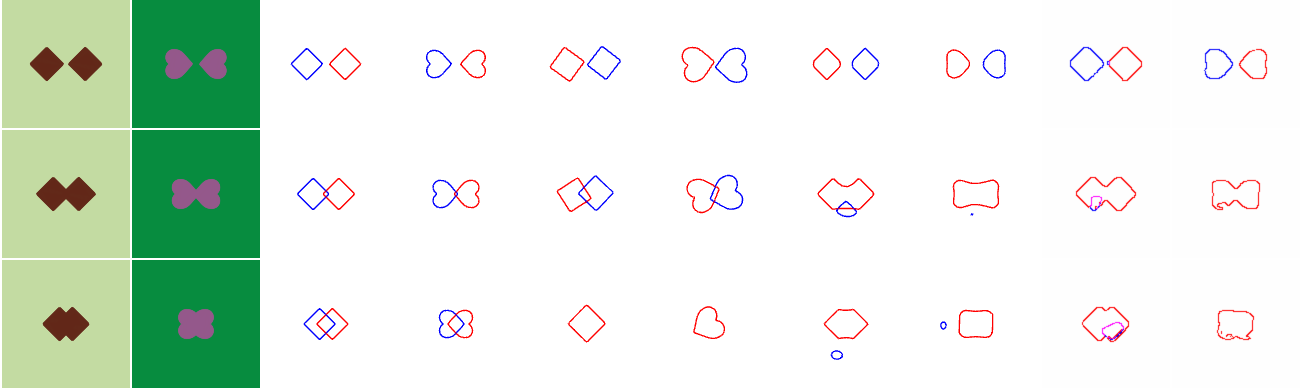}};
        \node[anchor=south, xshift=-0.4\textwidth] at (imageNode.north) {\footnotesize{Input images}};
        \node[anchor=south, xshift=-0.2\textwidth] at (imageNode.north) {\footnotesize{Ground truth}};
        \node[anchor=south, xshift=0.0\textwidth]  at (imageNode.north) {\footnotesize{\mname{TI-TP}}};
        \node[anchor=south, xshift=0.2\textwidth]  at (imageNode.north) {\footnotesize{Opt-Iter}};
        \node[anchor=south, xshift=0.4\textwidth]  at (imageNode.north) {\footnotesize{MONet}};
    \end{tikzpicture}
    \caption{Impact of object overlap on shape recovery.}
    \label{fig:overlap}
\end{figure}

\textbf{Overlapping objects.}
Figure \ref{fig:overlap} juxtaposes  \mname{TI-TP} and the baselines in terms of their capacity to recover the shapes of partially overlapping objects. Our model is capable of delineating the objects for moderate degree of overlap (middle row), which is prohibitively difficult for the baselines. When the overlap is high (bottom row), all methods fail. For \mname{TI-TP}, this is due to the DETR component segmenting the percept as a single object.  

\section{Limitations and future work}\label{sec:limitations}

The above observations are based on a single, synthetic benchmark. However, MDS-HR features a diversified repertoire of shapes, colors, and other visual aspects, providing therefore a representative sample of visual stimuli, similarly to the popular Multi-dSprites \cite{Burgess_2019}. In an ongoing follow-up work, we are applying \mname to medical imaging (histopathology and cytopathology), attempting to model the visual characteristics of individual cells in human tissues.  

\mname is limited to interpretation of 2D scenes and makes strong assumptions about object appearance (uniform coloring, no textures) and their interactions (simple blending mode when object overlap). However, the main point of this study was to devise (i) a general blueprint for gradient-trainable scene understanding models that involve elements of physical plausibility and (ii) ways in which such models can be effectively trained. We hypothesize (and plan to investigate in future work) that most of those limitations can be addressed by `lifting' the corresponding elements of our approach. For instance, \mname could be relatively easily generalized to 3D scenes by replacing the current shape model with 3D models, updating the definitions of visual aspects to 3D, and equipping it with a 3D renderer.

\bibliographystyle{alpha}
\bibliography{bibliography}

\newpage
\appendix


\section{The MDS-HR dataset}\label{sec:mds-hr}

\begin{figure}[b!]
    \centering
    \begin{tikzpicture}[
        node distance = 1pt,
        every node/.style = {inner sep=0pt}
    ]
    
      \node (pic1) {\includegraphics[width=0.85\linewidth]{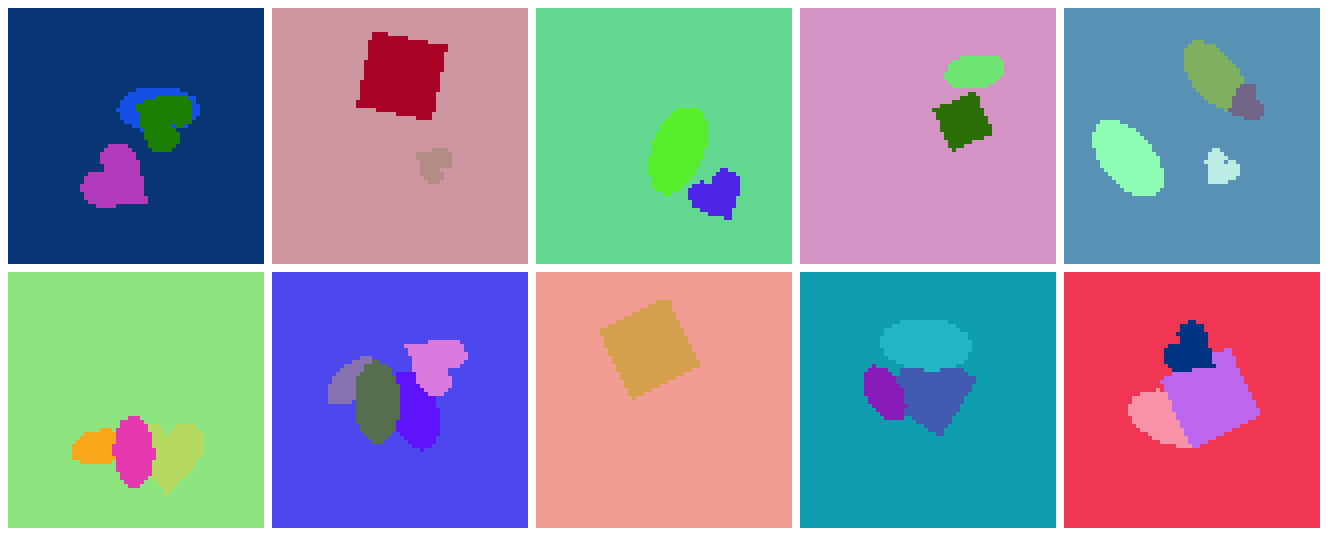}};
      \node[left=1mm of pic1, font=\footnotesize, align=center] {multi-\\dSprites};
    
      \node (pic2)[below=10pt of pic1] {\includegraphics[width=0.85\linewidth]{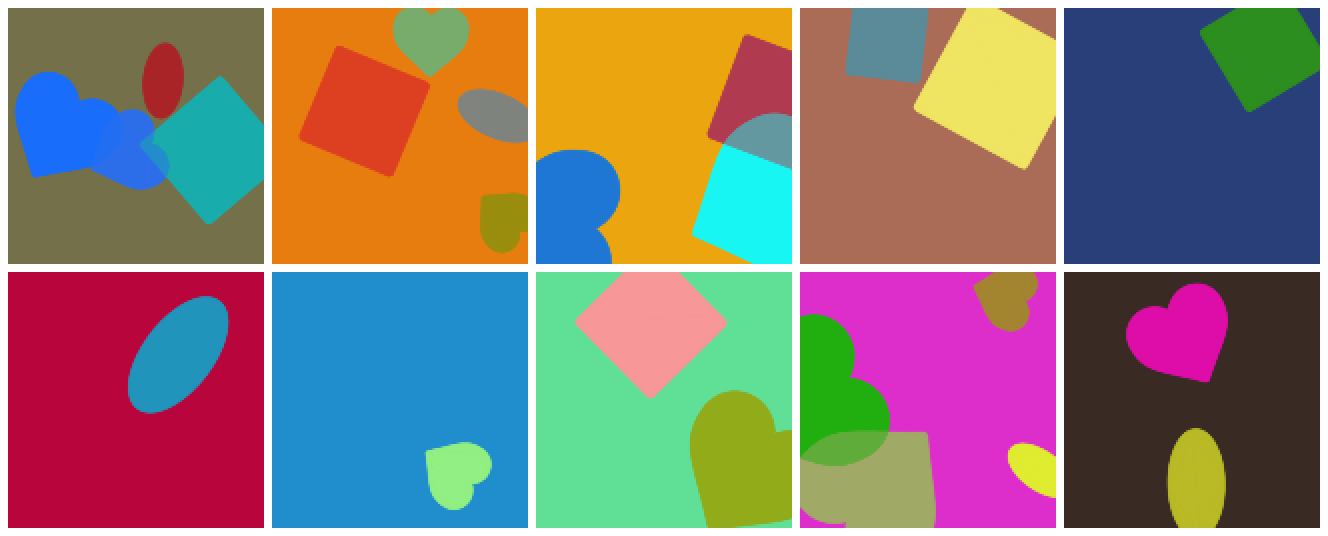}};
      \node[left=1mm of pic2, font=\footnotesize, align=center] {MDS-HR};
    
    \end{tikzpicture}
    \caption{ The first ten examples from the training sets of Multi-dSprites (top) and MDS-HR (bottom).  Best viewed when zoomed in. }
    \label{fig:mds-vs-mds-hr}
\end{figure}

The benchmark proposed and used in this paper, dubbed MDS-HR, is inspired by the popular Multi-dSprites benchmark \cite{Burgess_2019}, but diverges from it in several respects. The primary motivation for its development was the low resolution of Multi-dSprites (64x64), making the disentanglement task more challenging, 
because the pre-trained models serving as feature extractors in \mname (DINOv2 \cite{oquab2023dinov2} in our case) struggle to provide useful features for robust characterization of objects comprising just a handful of pixels; see examples in the top row of Fig.\ \ref{fig:mds-vs-mds-hr}. To more accurately represent object boundaries in MDS-HR, we incorporated anti-aliasing, replacing the hard, jagged contours of Multi-dSprites. This better reflects the continuous nature of objects contours, which  in real-world images contain pixels that have intermediate values between the foreground and the background, rather than just having one of those colors.
The other important difference is that MDS-HR does not feature object occlusion. Instead, the colors of overlapping objects are blended. Moreover, in MDS-HR, objects are on average larger and can also extend beyond the viewport.
Images shown in in Figure \ref{fig:mds-vs-mds-hr} exemplify the differences between the datasets.

We generate the MDS-HR scenes by sampling several object parameters. The number of objects per scene is sampled uniformly from the interval $[1,4]$. Continuous parameters: color, rotation angle, translation, and scale, are sampled from uniform distributions over the ranges $[0, 1]$, $[0, 2\pi]$, $[0.05, 0.95]$, and $[0.1, 0.3]$\footnote{ A centered square with a scale of $0.5$ would completely cover the entire scene. }, respectively. To prevent excessive overlap, object positions are determined by sampling multiple sets of translation vectors for the objects (eight sets per object) and selecting the one that maximizes the minimum pairwise distance between the objects. Finally, object shapes are chosen uniformly from a predefined set: ellipse, heart, and square.

The source code used to generate MDS-HR is available as a part of the code based described in Section \ref{sec:software}, Running the scripts available there with the above default parameterization produces MDS-HR as used in this paper, comprising the training and testing sets of, respectively \num{55000} and \num{5000} examples. Obviously, it is possible to generate auxiliary sets by changing the above parameters and/or the seed of the random number generator.


\section{Representing shapes with Elliptic Fourier Descriptors (EFD)}\label{sec:efd}

All spectral representations of shape start with the observation that a closed contour of a 2D object, represented as a sequence of $K$ points $(x_p,y_p)$ such that $x_1 = x_K$ and $y_1 = y_K$, can be alternatively viewed as a 1D cyclic complex time series $x_p+jy_p$, where $j$ is the imaginary unit. Such time series lend themselves naturally to discrete Fourier transform, with the coefficients of the resulting spectrum conveniently capturing the coarse aspect of the shape in the low frequencies and the details in the higher frequencies. Another advantage of this class of shape representations is the existence of the inverse transform, which facilitates shape processing and synthesis. 

The elliptic Fourier Transform (EFD) \cite{kuhl1982} is a particular method from this family. In contrast to the more basic representation of shape with simple Fourier descriptors, which relies on polar coordinates and assumes the ray length to be a \emph{function} of the angle, EFD can model arbitrary shapes, including such that require the contour to intersect itself. The EFD coefficients order $N$ are defined as:
\begin{equation}
\label{eqn:efd}
\begin{gathered}
A_n = \frac{T}{2 n^2 \pi^2} \sum_{p=1}^{K} \frac{\Delta x_p}{\Delta t_p} \left(\cos\frac{2n\pi t_p}{T} - \cos\frac{2n\pi t_{p-1}}{T} \right)  \\
B_n = \frac{T}{2 n^2 \pi^2} \sum_{p=1}^{K} \frac{\Delta x_p}{\Delta t_p} \left(\sin\frac{2n\pi t_p}{T} - \sin\frac{2n\pi t_{p-1}}{T} \right)  \\
C_n = \frac{T}{2 n^2 \pi^2} \sum_{p=1}^{K} \frac{\Delta y_p}{\Delta t_p} \left(\cos\frac{2n\pi t_p}{T} - \cos\frac{2n\pi t_{p-1}}{T} \right) \\
D_n = \frac{T}{2 n^2 \pi^2} \sum_{p=1}^{K} \frac{\Delta y_p}{\Delta t_p} \left(\sin\frac{2n\pi t_p}{T} - \sin\frac{2n\pi t_{p-1}}{T} \right) \\
\end{gathered}
\text{ for } n = 1,2,...,N
\end{equation}
\noindent where 
$\Delta x_p \equiv x_p - x_{p-1}$, $\Delta y_p \equiv y_p - y_{p-1}$, $\Delta t_p = \sqrt{\Delta x_p^2 + \Delta y_p^2}$ and $T = \sum_{p=1}^{K}\Delta t_p$. 
\vspace{3mm}
The coefficients $A_n$, $B_n$, $C_n$, and $D_n$ form together the EFD. They are translation invariant by design and can be further normalized to be invariant w.r.t. rotation and scale.

The original contour can be reconstructed using the inverse transform given by the following equations:
\begin{equation}
\begin{gathered}
x_p = \sum_{n=1}^{N} \left(\ A_n \cos\frac{2n\pi t_p}{T} + B_n \sin\frac{2n\pi t_p}{T} \right) \\
y_p = \sum_{n=1}^{N} \left(\ C_n \cos\frac{2n\pi t_p}{T} + D_n \sin\frac{2n\pi t_p}{T} \right) \\
\end{gathered}
\text{ for } p = 1,2,...,K
\end{equation}

In \mname, the coefficients of EFD form the shape representation. We used $N=16$ and $K=64$. The $shape()$ component in the decoder head (Fig.\ \ref{fig:diagram}) maintains a bank of $k=3$ EFDs defined in this way, each represented as a vector, and producing a sequence of $(x_p,y_p)$s via the above inverse transform. $k$ is set to 3 by the clustering algorithm (see the final paragraphs of Sec.\ \ref{sec:training}), which successfully discovers that MDS-HR features three classes of shape. When queried on specific latent $z$, $shape(z)$ performs $k$-ary soft-indexing of these coordinates, and returns a weighted sum of them, to be passed to the renderer. As the inverse transform is differentiable, the gradient back-propagated from the renderer can effectively update the parameters of $shape()$ component, which learns how to pick the appropriate shape using soft indexing. The $k$ vectors representing the EFDs are not trainable themselves and remain fixed during training, as they have been estimated using the clustering-based algorithm described at the end of Sec.\ \ref{sec:training}.   

The parameter-space loss function $L_p$ (Eq.\ \ref{eq:L_p} and Sec.\ \ref{sec:L_p}), used in the GP and GI training modes (Sec.\ \ref{sec:training}) contains a term that measures the $L_2$ discrepancy in the contour space between the predicted EFD, $\hat{p}_i.sh$, with the target EFD, $p_i.sh$.


\section{The parameter-space loss function $L_p$}\label{sec:L_p}

The parameter space loss function introduced in Sec.\ \ref{sec:loss} comprises an object matching phase and a loss calculation phase. 
\begin{enumerate}
    \item 
In the matching phase, the first two terms of the per-object matching cost function, $C(p_i, \hat{p}_j)$ (Eq. \ref{eq:L_m}), aim to select the optimal match between object candidates and ground truth objects on translation and color. The third term enforces matching stability, as we observed that instability (i.e., the matching process selecting an object with low \emph{confidence} -- not preferred by the model) hindered the training process.  In training, examples are grouped by the number of ground truth objects to form chunks inside the batches, enabling vectorized computation on these chunks. We utilized the Hungarian method implementation available in the SciPy\footnote{\url{https://scipy.org/}} library due to the lack of an efficient GPU-based counterpart.

\item
During loss calculation, determining whether a candidate is a part of the scene is formulated as a binary classification task. Matched candidates serve as the positive class, unmatched candidates as the negative class, and Binary Cross-Entropy (BCE) is used for this loss (Eq. \ref{eq:L_p}, lines 2 and 8). A significantly larger weight on the translation term (Eq. \ref{eq:L_p}, line 3) is a form of \emph{curriculum learning} -- as we found out in preliminary experiments, \mname models struggle to predict any of the object parameters until they have learned to localize  objects. 

The function $sym$ present in the term focused on rotation computes the number of symmetries for shapes represented by Elliptical Fourier Descriptors (EFDs). It first identifies \emph{significant} EFD harmonics by comparing their magnitudes ($L2$ norms of coefficients) against a threshold. The core of the method lies in calculating the Greatest Common Divisor (GCD) of the indices of these significant harmonics. This GCD value is returned as the order of rotational symmetry, defaulting to 1 if no harmonics are  sufficiently significant. This handling of the prediction angle rewards the model with low loss values if it predicts the rotation angle correctly \emph{modulo the number of object's symmetries}, or, more precisely, modulo $2\pi/sym()$ (e.g., modulo 90 angular degrees for squares). 
\end{enumerate}


\section{Training of models}\label{sec:training-details}

All models were trained using the Adam optimizer \cite{DBLP:journals/corr/KingmaB14} with a learning rate of 0.0001 and a budget of 100 epochs. 
For training, all \mname configurations used a batch size of 128 (the largest power of two size fitting into the GPU memory), while MONet used a batch size of 64 (due to the same constraint). For testing, models were selected based on their $L_x$ scores in the validation set. The validation set contained \num{5000} examples and has been sampled randomly from the training set, effectively reducing the size of the latter from \num{55000} to \num{50000}. The validation set contains the same examples in all configurations examined in experiments. 


\section{Additional results}\label{sec:extra-results}

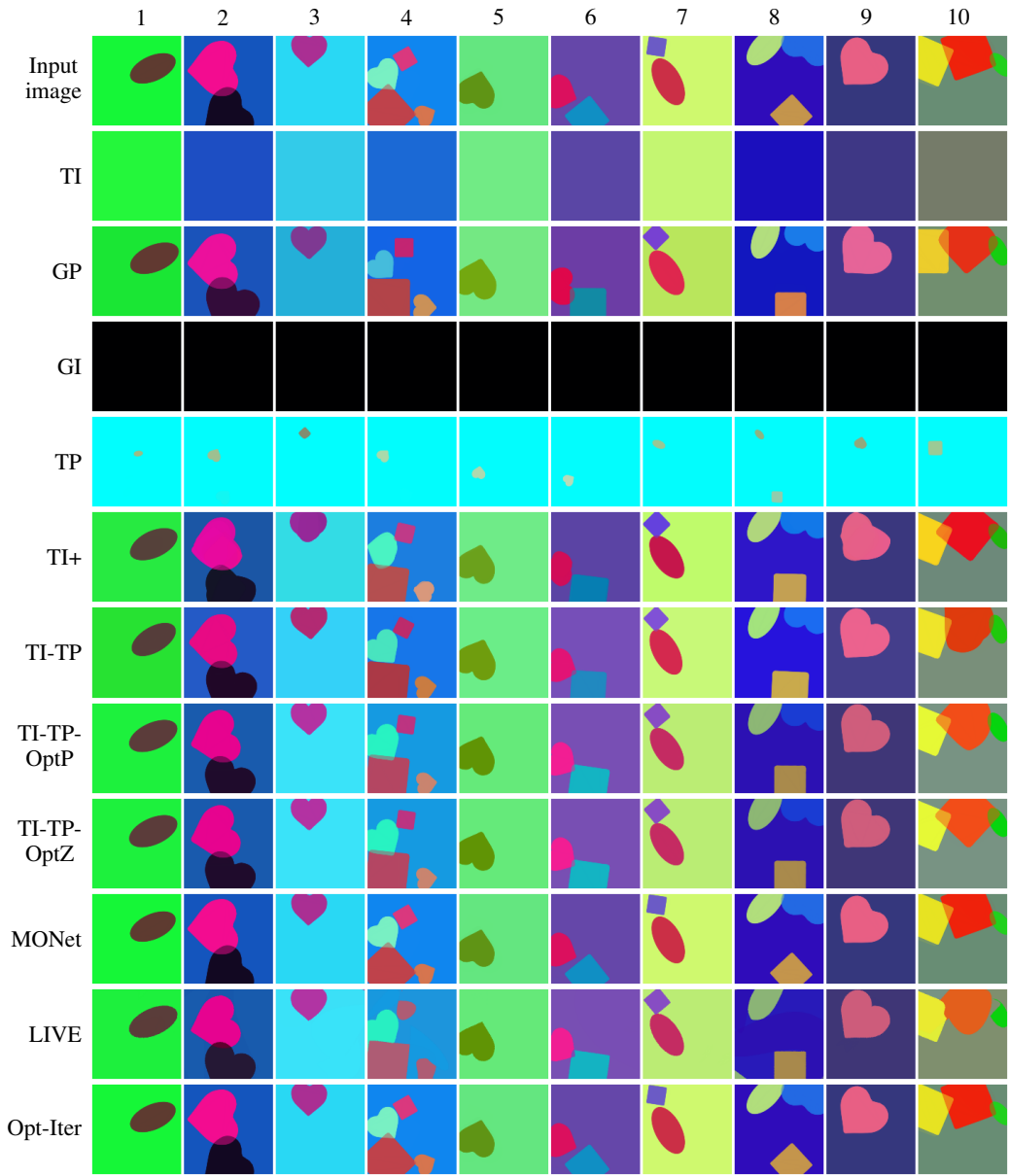
\begin{figure}
    \centering
    \begin{tikzpicture}[
        node distance = 1pt,
        every node/.style = {inner sep=0pt}
    ]
    
      \node (GT) {\includegraphics[width=0.9\linewidth]{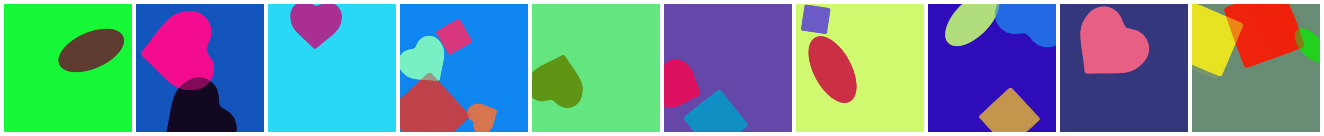}};
      \node[left=1mm of GT, font=\footnotesize, align=center] {Input\\image};

      \node (TI)[below=0pt of GT] {\includegraphics[width=0.9\linewidth]{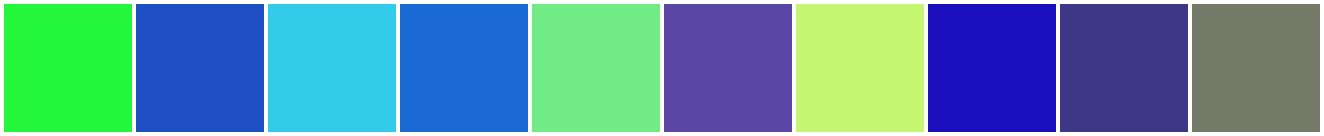}};
      \node[left=1mm of TI, font=\footnotesize, align=center] {TI};

      \node (GP)[below=0pt of TI] {\includegraphics[width=0.9\linewidth]{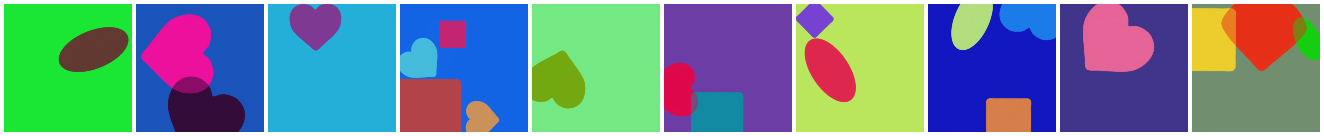}};
      \node[left=1mm of GP, font=\footnotesize, align=center] {GP};

      \node (GI)[below=0pt of GP] {\includegraphics[width=0.9\linewidth]{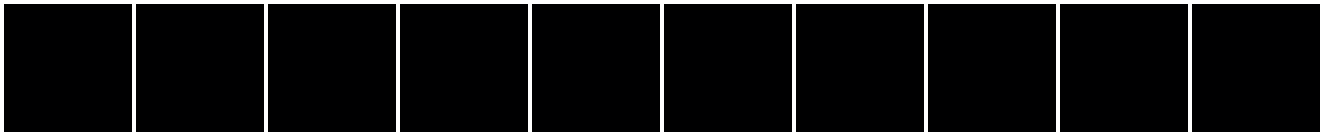}};
      \node[left=1mm of GI, font=\footnotesize, align=center] {GI};

      \node (TP)[below=0pt of GI] {\includegraphics[width=0.9\linewidth]{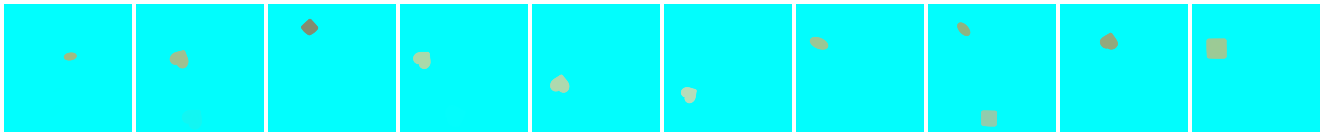}};
      \node[left=1mm of TP, font=\footnotesize, align=center] {TP};

      \node (TI+)[below=0pt of TP] {\includegraphics[width=0.9\linewidth]{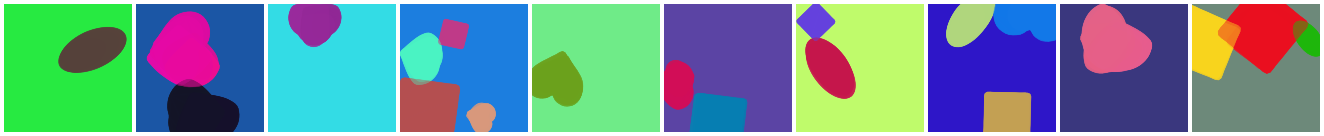}};
      \node[left=1mm of TI+, font=\footnotesize, align=center] {TI+};

      \node (TI-TP)[below=0pt of TI+] {\includegraphics[width=0.9\linewidth]{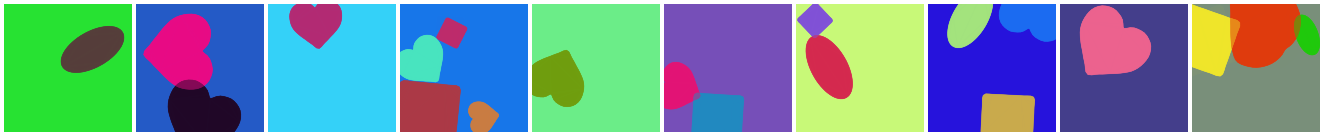}};
      \node[left=1mm of TI-TP, font=\footnotesize, align=center] {TI-TP};

      \node (TI-TP-OptP)[below=0pt of TI-TP] {\includegraphics[width=0.9\linewidth]{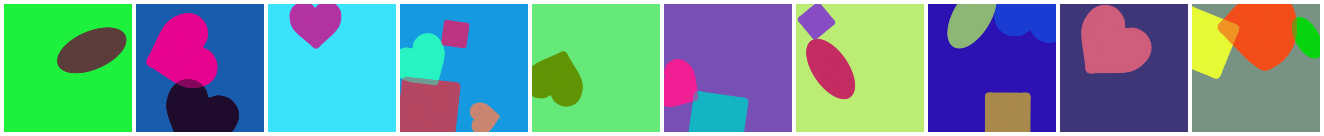}};
      \node[left=1mm of TI-TP-OptP, font=\footnotesize, align=center] {TI-TP-\\OptP};
      
      \node (TI-TP-OptZ)[below=0pt of TI-TP-OptP] {\includegraphics[width=0.9\linewidth]{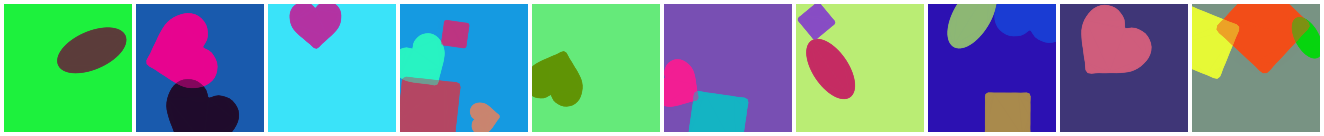}};
      \node[left=1mm of TI-TP-OptZ, font=\footnotesize, align=center] {TI-TP-\\OptZ};

      \node (MONet)[below=0pt of TI-TP-OptZ] {\includegraphics[width=0.9\linewidth]{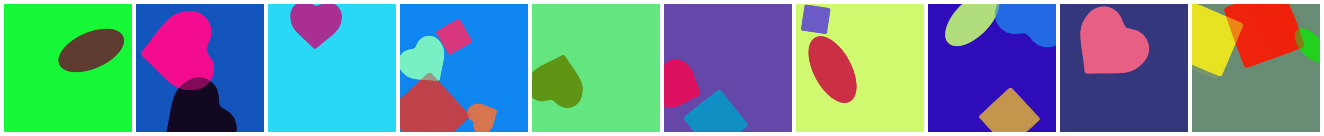}};
      \node[left=1mm of MONet, font=\footnotesize, align=center] {MONet};

      \node (LIVE)[below=0pt of MONet] {\includegraphics[width=0.9\linewidth]{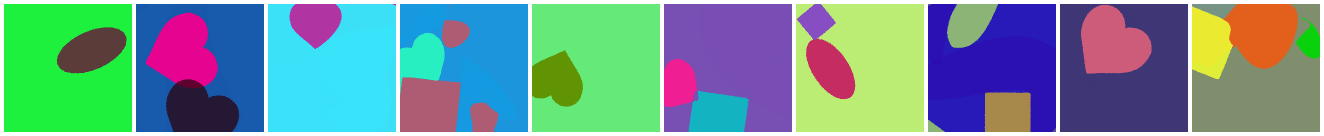}};
      \node[left=1mm of LIVE, font=\footnotesize, align=center] {LIVE};

      \node (Opt-Iter)[below=0pt of LIVE] {\includegraphics[width=0.9\linewidth]{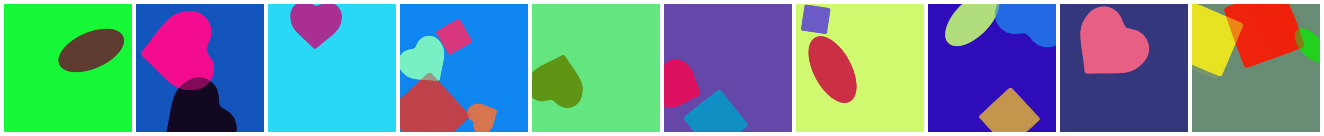}};
      \node[left=1mm of Opt-Iter, font=\footnotesize, align=center] {Opt-Iter};

\node[anchor=south, xshift=-0.4 \textwidth, yshift=1mm, font=\footnotesize] at (GT.north) {1};
\node[anchor=south, xshift=-0.32\textwidth, yshift=1mm, font=\footnotesize] at (GT.north) {2};
\node[anchor=south, xshift=-0.23\textwidth, yshift=1mm, font=\footnotesize] at (GT.north) {3};
\node[anchor=south, xshift=-0.14\textwidth, yshift=1mm, font=\footnotesize] at (GT.north) {4};
\node[anchor=south, xshift=-0.05\textwidth, yshift=1mm, font=\footnotesize] at (GT.north) {5};
\node[anchor=south, xshift= 0.04\textwidth, yshift=1mm, font=\footnotesize] at (GT.north) {6};
\node[anchor=south, xshift= 0.13\textwidth, yshift=1mm, font=\footnotesize] at (GT.north) {7};
\node[anchor=south, xshift= 0.22\textwidth, yshift=1mm, font=\footnotesize] at (GT.north) {8};
\node[anchor=south, xshift= 0.31\textwidth, yshift=1mm, font=\footnotesize] at (GT.north) {9};
\node[anchor=south, xshift= 0.4 \textwidth, yshift=1mm, font=\footnotesize] at (GT.north) {10};

    \end{tikzpicture}
    \caption{ More reconstruction results for selected images from the testing set (cf. Fig.\ \ref{fig:mds-hr-ex}). }
    \label{fig:mds-hr-more}
\end{figure}

Figure \ref{fig:mds-hr-more} presents more examples of scene reconstructions produced by the methods compared in Sec.\ \ref{sec:experiment}, presented similarly to Fig.\ \ref{fig:mds-hr-ex}.  Most of the basic configurations of \mname (TI, GP, GI and TP) trained in the simple training regimes, i.e. with one of the considered loss functions (in the image space or in the parameter space) fail almost completely, for the reasons discussed in Sec.\ \ref{sec:experiment}. The only exception is GP, the variant trained with the parameter-space loss function $L_p$. In contrast, all combined variants of \mname and the baselines reproduce images quite well, and in some cases almost perfectly. Recall, however, that for MONet it is a given, as it represents objects with raster masks, so it does not inherently model the notion of shape. 

Bar the trivial omissions of objects by the basic variants of \mname, the visual aspect that seems to be most problematic for the method is rotation. Translation, scale and colors (both background and object) do not pose significant challenges. This corroborates the conclusions of our discussion on the particular strong entanglement of this aspect with object shape, which motivated us to detach the learning of shapes from the rest of the training process (see the final paragraphs of Sec.\ \ref{sec:training}).

Notice that most methods and configurations struggle with objects cropped by the boundaries of the viewport (except for MONet, for the reasons indicated above). This is particularly true for rectangles (e.g. the red one in \#4, the blue one in \#6, the orange one in \#8), probably due to this shape having the highest number of symmetries among the shapes present in MDS-HR (cf. the role of the number of symmetries in the discussion on the parameter-space loss function $L_p$ in Sec.\ \ref{sec:L_p}). This happens less often for the heart shape and mostly for the more basic variants of the approach (e.g. \#6 for GP and TI).

\section{Software implementation}\label{sec:software}

Concerning the \textbf{overall architecture} of \mname, we implemented it in PyTorch\footnotemark{}, with particular support of the following libraries:

\footnotetext{Adam Paszke, Sam Gross, Soumith Chintala, Gregory Chanan, Edward Yang,
Zachary DeVito, Zeming Lin, Alban Desmaison, Luca Antiga, and Adam Lerer. Automatic differentiation in PyTorch. \textit{In NIPS-W}, 2017.}

\begin{itemize}
    \item Torchvision\footnote{\url{https://github.com/pytorch/vision}}
    \item TorchMetrics\footnote{\url{https://github.com/Lightning-AI/torchmetrics}}
    \item TensorDict\footnote{\url{https://github.com/pytorch/tensordict}}
    \item Kornia\footnote{\url{https://github.com/kornia/kornia}}
    \item scikit-learn\footnote{\url{https://scikit-learn.org/}}
    \item torchtyping\footnote{\url{https://github.com/patrick-kidger/torchtyping}}
    \item typeguard\footnote{\url{https://github.com/agronholm/typeguard}}
\end{itemize}
The constituent models and components came from the following sources:
\begin{itemize}
    \item DINO implementation and weights from the PyTorch Hub\footnote{\url{https://pytorch.org/hub/}}
    \item Conditional DETR\footnotemark{} (a DETR-derived architecture) implementation from HuggingFace's Transformers library\footnote{\url{https://github.com/huggingface/transformers}} 
\end{itemize}

\footnotetext{Depu Meng, Xiaokang Chen, Zejia Fan, Gang Zeng, Houqiang Li, Yuhui Yuan, Lei Sun, and Jingdong Wang. Conditional DETR for fast training convergence. \textit{In Proceedings of the IEEE International Conference on Computer Vision (ICCV)}, 2021.} 

The source code of the implementation is available as an anonymous Git repository under the following location: 
\url{https://anonymous.4open.science/r/dvp_plus-8E1E}
We made it available under the permissive MIT license. The repository contains also the functions and scripts used to populate the MDS-HR benchmark (Sec.\ \ref{sec:mds-hr}) and the actual benchmark as well. The README.md file in the root folder of the repository contains the instructions on preparation of the Python environment and installation, and points to Jypyter notebooks for training and querying of particular variants of \mname as well as the baseline methods. 

For the \textbf{differentiable rendering} we used the implementation available in the PyTorch3D\footnote{\url{https://pytorch3d.org/}} \cite{ravi2020pytorch3d}. PyTorch3D's renderer is based on the SoftRas method \cite{softras}. While PyTorch3D offers various utilities for handling many types of shape representations, integrating it with our model required implementing additional steps. The resulting rendering pipeline comprises the following steps: 
\begin{enumerate}
\item Our model predicts a shape represented as a vector of Elliptical Fourier Descriptors (EFDs; Sec.\ \ref{sec:efd}), 
\item The EFD is transformed into a shape contour -- a sequence of $(x_p, y_p)$, 
\item The sequence is subject to the geometric transformations (translation, rotation, scaling) determined by the predictors in the decoder head (Fig.\ \ref{fig:diagram}) of the remaining visual aspects (see Sec.\ \ref{sec:arch} for the list of aspects). 
\item The transformed contour is triangularized. This step generates a list of indices that point to the contour's vertices to form a mesh of triangles defining the shape (conceptually similar to the index buffer used in indexed drawing in computer graphics). For this task, we implemented PyTorch bindings for the earcut.hpp\footnote{\url{https://github.com/mapbox/earcut.hpp}} library. 
\item Finally, the array of indices and the array of vertices, containing position, color, and confidence (as predicted by the encoder head; Fig.\ \ref{fig:diagram}), assigned to each vertex, are passed to the renderer.
\end{enumerate}

Concerning the actual rendering algorithm, inspired by the SoftRas method \cite{softras}, we compute a signed distance function $d(i, j)$ for each pixel $i$ and triangle $j$. This rasterization step is accelerated using CUDA kernels in PyTorch3D. A sigmoid function then transforms these distances into a soft mask $D^i_j$ representing the influence of each triangle on each pixel: 
\begin{equation}
D^i_j = \sigmoid\left(\frac{d(i, j)}{\sigma}\right)
\end{equation}
These masks, combined with an object-specific confidence score $f_j$ (for the object containing triangle $j$),  predicted in the decoder head, determine the final per-pixel, per-triangle weights $w^i_j$. 
\begin{equation}
w^i_j = \frac{D^i_j \exp\left(f_j / \gamma\right)}{\sum_k D^i_k \exp\left(f_k / \gamma\right)}
\end{equation}
Finally, the resulting image $I$ is calculated as a weighted sum of colors (where $C_j$ is the color of $j$th triangle and $C_b$ is the background color):
\begin{equation}
\begin{split}
\alpha &= 1 - \prod_j \left(1 - w^i_j C_j\right) \\
I^i &= \alpha\sum_j w^i_j C_j + (1 - \alpha) C_b
\end{split}
\end{equation}


\section{Hardware specification}\label{sec:hardware}

The configuration of the workstation used in the experimental part of the study was quite moderate:
\begin{itemize}
    \item CPU: Intel Core i7-11700KF (3.6 GHz base, up to 5.0 GHz boost, 8 cores, 16 threads)
    \item GPU: NVIDIA GeForce RTX 3090 (24 GB VRAM)
    \item RAM: 32 GB DDR4 (3200 MT/s)
    \item Storage: 2 TB Kingston KC3000 NVMe
    \item Operating system: Ubuntu 22.04.5 (via WSL 2 on Windows 11)
\end{itemize}

Training of a single \mname model on this hardware does not exceed 10 hours, with the actual times that we observed in our experiments ranging from a minimum of 6 hours 49 minutes for \mname{TI+} to a maximum of 9 hours 21 minutes for \mname{TI-TP}. 


\section{Experimental configurations}\label{sec:configs}

The \mname architecture, as described in Sec. \ref{sec:arch}, has the following configuration:

\begin{description}
\item{Encoder}
  \begin{enumerate}
    \item Pre-trained vision transformer: DINOv2 (\texttt{dinov2{\_}vitb14{\_}reg} variant) was used as the feature extractor. Its weights were kept frozen during all \mname training.
    \item Encoder-Decoder Transformer (EDT): We used a standard Conditional DETR\footnotemark{}  transformer (6 encoder layers, 6 decoder layers, 8 attention heads, 256 hidden dimensions) for the EDT component. We had to disable the dropout, as it hindered training.
    \item Object queries: 8 learnable object queries were used to probe the EDT.
  \end{enumerate}
\item{Decoder} \\ The decoder head consists of seven independent Multi-Layer Perceptrons (MLPs). Each MLP has 3 hidden layers of 256 units each, and ReLU activations between them. The MLPs vary only in the size of the output layer (determined by the dimensionality of the predicted vector, e.g. 3 for color) and in the activation function:
  \begin{enumerate}
    \item Color, translation, scaling, confidence, shape, and background color MLPs use the sigmoid as the final activation.
    \item Rotation MLP used tanh as the final activation, followed by a normalization to unit length.
  \end{enumerate}
\end{description}

\footnotetext{Depu Meng, Xiaokang Chen, Zejia Fan, Gang Zeng, Houqiang Li, Yuhui Yuan, Lei Sun, and Jingdong Wang. Conditional DETR for fast training convergence. \textit{In Proceedings of the IEEE International Conference on Computer Vision (ICCV)}, 2021.} 

For the \mname{TI-TP-OptP} and \mname{TI-TP-OptZ}, 100 iterations of the Adam optimizer with the learning rate of 0.01 were used for the per-scene optimization of, respectively, parameters $p$ or latents $z$. The learning rate was controlled by PyTorch's \texttt{ReduceLROnPlateau} learning rate scheduler configured with \texttt{patience} of 10 iterations,  \texttt{cooldown} of 10 iterations, and \texttt{factor} of 0.5.


\section{Baseline methods}\label{sec:baselines}

\textbf{LIVE.}
As signaled in the main text, LIVE \cite{live} operates similarly to our Opt-Iter baseline (because our Opt-Iter was inspired and based on LIVE). LIVE iteratively identifies the image region that is most divergent from the current scene approximation, initializes a new object candidate at that location (sampling its color from the region), and then optimizes all candidates introduced so far to minimize $L_x$. To represent object shape, LIVE uses closed Bézier paths (in contrast to Opt-Iter, which relies on EFDs -- as a matter of fact, this is the main difference between both methods). 
The cycle repeats until a predetermined number of objects is reached.  
By adding consecutive object candidates sequentially, LIVE attempts to address the risk of single objects being reconstructed with multiple object candidates. This also leads to reconstructing the target with fewer primitives, making the reconstruction more interpretable.

We used the official LIVE implementation\footnote{\url{https://github.com/Picsart-AI-Research/LIVE-Layerwise-Image-Vectorization}}. The results reported in Sec.\ \ref{sec:experiment} employed the predefined \texttt{experiment{\_}5x1} configuration.

\textbf{MONet.} 
The Multi-Object Network (MONet) \cite{Burgess_2019} is a composite architecture that combines image segmentation based on an attention mechanism (to delineate image components) with a variational autoencoder (VAE), for rendering individual components in the scene. Similarly to our approach, it attempts to reconstruct the input image/scene and can be trained end-to-end with gradient. However, MONet does not involve geometric aspects of image formation, in particular it does not perform geometric rendering of objects: the subimages of individual components are generated with the VAE and `inpainted' into the scene using raster masks.   

Our implementation of MONet is based on Object-Centric Library\footnote{\url{https://github.com/addtt/object-centric-library}} and uses the following parameter setting: 5 slots and 6 U-Net blocks.


\end{document}